\documentclass[11pt]{article}

\usepackage{dialogue}

\usepackage{times}
\usepackage{latexsym}
\usepackage{amsmath}

\usepackage{subfigure}
\usepackage{graphicx}
\usepackage{ifxetex}
\usepackage{ifxetex}
\ifxetex
\usepackage{fontspec}
\setromanfont{Times New Roman}
\else
\usepackage{cmap}
\usepackage[T2A,T1]{fontenc}
\usepackage[utf8]{inputenc}
\usepackage{times}
\usepackage{latexsym}
\usepackage{substitutefont}
\substitutefont{T2A}{\familydefault}{cmr}
\fi

\usepackage[russian,british]{babel}
\usepackage{url}
\usepackage{pgf}
\usepackage{makecell}
\usepackage{covington} 

\usepackage{hyperref}

\dialogfinalcopy 
\title{Artificial Text Detection with Multiple Training Strategies}
\author{Bin Li\textsuperscript{1}\thanks{ \ \ These authors contribute equally to this work.},   Yixuan Weng\textsuperscript{2$*$}, Qiya Song\textsuperscript{1$*$}, Hanjun Deng\textsuperscript{3}\\
	$^1$ 
	College of Electrical and Information Engineering, Hunan University, Changsha, China
	\\
	$^2$ National Laboratory of Pattern Recognition Institute of Automation, 
	\\Chinese Academy Sciences, Beijing, China
	\\
	$^3$ Experimental High School Affiliated to Beijing Normal University, Beijing, China
	\\
	\{libincn, sqyunb\}@hnu.edu.cn, wengsyx@gmail.com, Hanjun\_Deng@gmail.com
}
\begin{document}
	\maketitle
	\begin{abstract}
		As the deep learning rapidly promote, the artificial texts created by generative models are commonly used in news and social media. However, such models can be abused to generate product reviews, fake news, and even fake political content. The paper proposes a solution for the Russian Artificial Text Detection in the Dialogue shared task 2022 (RuATD 2022) to distinguish which model within the list is used to generate this text. We introduce the DeBERTa pre-trained language model with multiple training strategies for this shared task. 
		Extensive experiments conducted on the RuATD dataset validate the effectiveness of our proposed method. Moreover, our submission ranked \textbf{second place} in the evaluation phase for RuATD 2022 (Multi-Class).
		
		
		\textbf{DOI:} 10.28995/2075-7182-2022-20-104 
	\end{abstract}
	\selectlanguage{russian}
	\begin{center}
		\russiantitle{Обнаружение искусственного текста с несколькими стратегиями обучения}
		
 		\medskip \setlength\tabcolsep{1.5em}
 		\begin{tabular}{cc}
 			\textbf{Бин Ли} & \textbf{Йи Суан Вен} 
 			\\
 			Хунаньский университет & Китайская академия наук 
 			\\
 			Район Юэлу, город Чанша & Район Хайдянь, Пекин
 			\\
 			{\tt libincn@hnu.edu.cn} &  {\tt wengsyx@gmail.com} 
 			\vspace{0.2cm}\\
			\textbf{Кийя Сон} & \textbf{Хан Джун Дэн} 

			\\
			Хунаньский университет & \makecell{Средняя школа при \\ Пекинском педагогическом университете}
			\\
			Район Юэлу, город Чанша &  Район Сичэн, Пекин
			\\
			{\tt sqyunb@hnu.edu.cn} &  {\tt Hanjun\_Deng@gmail.com} 
 		\end{tabular}
 		\medskip
	\end{center}
	
	\begin{abstract}
 С быстрым распространением глубокого обучения искусственный текст, созданный с помощью генеративных моделей, обычно используется в новостях и социальных сетях. Однако такими моделями можно злоупотреблять для создания обзоров продуктов, фейковых новостей и даже фальшивого политического контента. В документе предлагается решение задачи совместного использования диалога по обнаружению искусственного текста на русском языке 2022 (RuATD 2022), чтобы различать, какая модель в списке используется для создания искусственного текста. Мы представляем предварительно обученные языковые модели DeBERTa с несколькими стратегиями обучения, а обширные эксперименты с набором данных RuATD подтверждают эффективность предложенного нами метода. Кроме того, представленные нами результаты заняли \textbf{второе место} на этапе оценки RuATD 2022 (Мультиклассовая). 
		
\textbf{Ключевые слова:} Обнаружение искусственного текста, предварительно обученная языковая модель, несколько стратегий обучения.
	\end{abstract}
	\selectlanguage{british}
	
	\section{Introduction}
	\label{intro}
	With the rapid development of AI technologies, a growing number of methods the ability to generate realistic artifacts. For instance, amount of texts
	generated by recent text generation methods the transformer encoder-decoder framework are very close to the text written by humans, including lots of security issues \cite{de2021survey,topal2021exploring}. Extensive Transformer-based text generation models, such as Bert-stytle \cite{2018bert}, GPT-stytle \cite{2019language}, have achieved excellent results on a large number of NLP tasks. \cite{2019ctrl} proposed the conditional transformer language model (CTRL) with 1.63 billion parameters to control the text generation.  The model is trained with different codes that control  task-specific behavior, entities, specify style and content. \cite{2019defending} introduced a controllable text generation model named Grover, which can overwrite propaganda papers. For example, given a headline "Discovering the link between vaccines and autism," Grover could generate a description article for this title. Humans rated this generated text as more trustworthy than human-written text.
	However, the success of natural language generation has drawn dual-use concerns. On the one hand, applications such as summary generation and machine translation are positive. On the other hand, related techniques may also enable adversaries to generate neural fake news, targeted propaganda and even fake political content. Therefore, several researchers have made many attempts to develop artificial text detectors \cite{2020automatic}. \cite{2019release} used the pre-trained language model RoBERTa for the downstream text detection task and achieved the best performance in recognizing web pages generated by the GPT-2 model.  \cite{2021artificial} proposed a novel method based on Topological Data Analysis (TDA). 
	The interpretable topological features that can be derived
	from the attention map of any transformer-based language model are introduced for the task of artificial text detection. \cite{shamardina2022ruatd} 
	originated two tracks  on the RuATD 2022 Dialogue Shared task to solve the problem of automatic recognition of generated texts. In this paper, we adopt  DeBERTa method with multiple training strategies for the Russian artificial text detection in the Dialogue shared task 2022 (Multi-Class). More details about our system are introduced in the following sections.
	\section{Main Method}
	This section will elaborate on the main method for the Russian artificial text detection dialogue shared task, where we adopt the pre-trained model with multiple training strategies, such as adversarial training, child tuning, and intrust loss function.
	\begin{figure*}[h]
		\centering
		\includegraphics[scale=0.62]{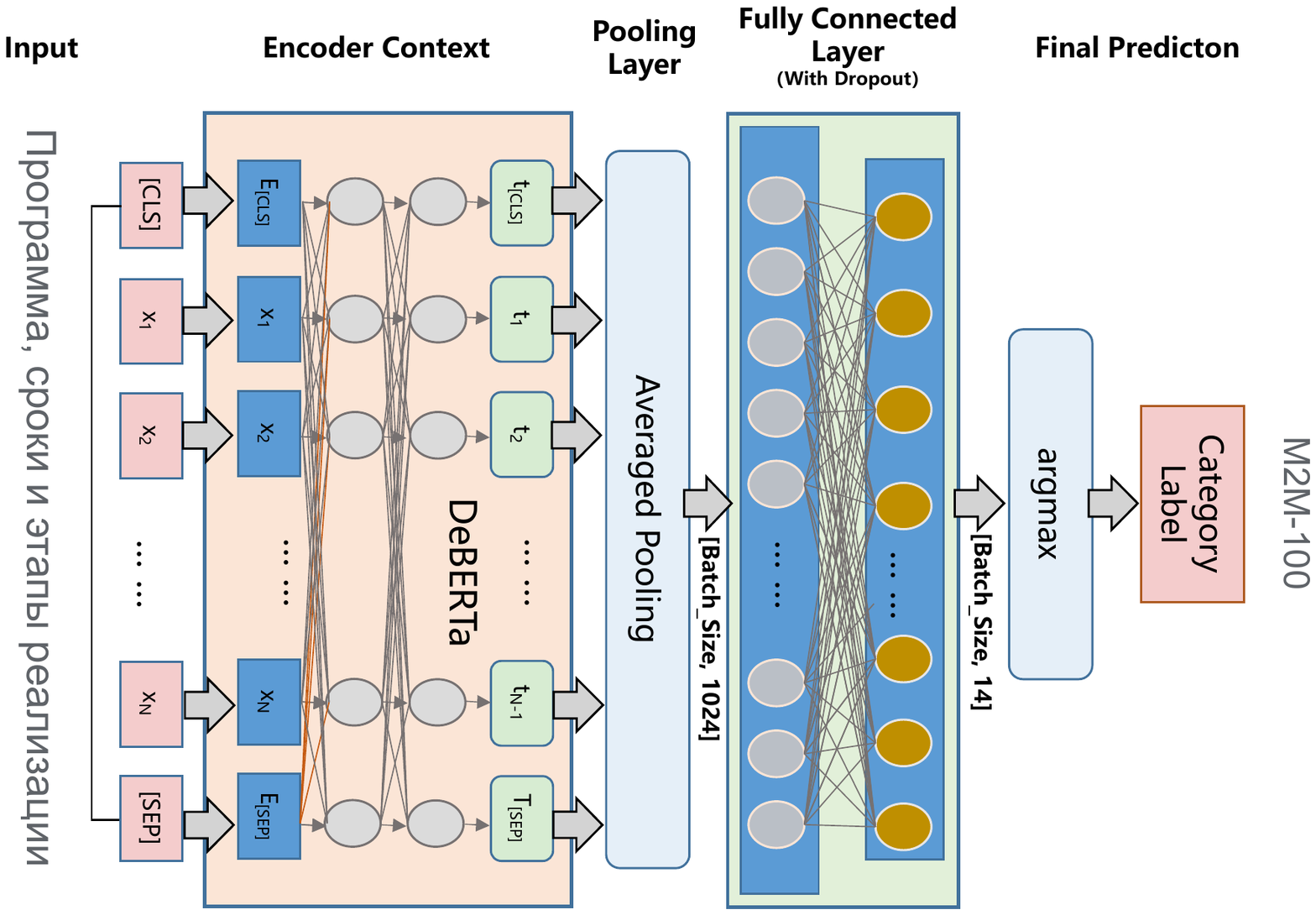}
		\caption{The model architecture of the submitted system.}
		\label{text21}
		\vspace{-1cm}
	\end{figure*}
	\subsection{Overview of Pre-trained Model}
	It is noted that the pre-trained model has a solid ability to differentiate the results generated by different models \cite{qiu2020pre,naseem2021comprehensive}, we resort to the state-of-the-art (SOTA) pre-trained language model for better prediction. As shown in Figure \ref{text21}, we present the main model architecture for the shared task. Specifically, we adopt the DeBERTa family, i.e., mDeBERTa \cite{he2020deberta} and DeBERTa \cite{he2021debertav3}, for the category classification. The pooling and fully connected layers are at the top of the pre-trained language model for leveraging global semantics. Finally, the argmax is performed after the 14 categories classification to obtain the final results.
	\subsection{Multiple Training Strategies}
	\subsubsection{Adversarial Training}
	The common method in adversarial training is the Fast Gradient Method \cite{nesterov2013gradient,dong2018boosting}. The idea of FGM is straightforward. Increasing the loss is to increase the gradient so that we can take 
	\begin{equation}
		\Delta x=\epsilon \nabla_{x} L(x, y ; \theta)
	\end{equation}
	where $x$ represents the input, $y$ represents the label, $\theta$ is the model parameter, $L(x,y;\theta)$ is the loss of a single sample, $\Delta x$ is the anti-disturbance.
	
	Of course, to prevent $\Delta x$ 
	from being too large, it is usually necessary to standardize 
	$\nabla_{x} L(x, y ; \theta)$. The more common way is
	\begin{equation}
		\Delta x=\epsilon \frac{\nabla_{x} L(x, y ; \theta)}{\left\|\nabla_{x} L(x, y ; \theta)\right\|} 
	\end{equation}
	\subsubsection{Child-tuning Training}
	The efficient Child-tuning \cite{xu2021raise} method is used to fine-tuning the backbone model in our method, where the parameters of the Child network are updated with the gradients mask. For this shared task, the task-independent algorithm is used for child-tuning. When fine-tuning, the gradient masks are obtained by Bernoulli Distribution \cite{chen1997statistical} sampling from in each step of iterative update, which is equivalent to randomly dividing a part of the network parameters when updating. The equation of the above steps is shown as follows 
	\begin{equation}
		\begin{aligned}
			&\mathit{w}_{t+1}=\mathit{w}_{t}-\eta \frac{\partial \mathcal{L}\left(\mathit{w}_{t}\right)}{\partial \mathit{w}_{t}} \odot B_{t} \\
			&\mathit{B}_{t} \sim \text { Bernoulli }\left(p_{F}\right)
		\end{aligned}
	\end{equation}
	where the notation $\odot$ represents the dot production, $p_{F}$ is the partial network parameter.
	\subsection{In-trust Loss}
	Followed the work \cite{huang2021named}, the noisy data sets may also provide knowledge. To reduce the entropy information between the noising synthetic and real labels given the text, we adopt the Incomplete-trust (In-trust) loss as the replacement for the original cross-entropy loss function, which is intended to train with uncertainty in the presence of noise. The new loss function is shown as follows
	\begin{equation}
		\begin{aligned}
			&L_{DCE}=-p \log (\delta p+(1-\delta) q) \\
			&L_{I n-t r u s t}=\alpha L_{C E}+\beta L_{D C E}
		\end{aligned}
	\end{equation}
	where $L_{DCE}$ is an acceleration adjustment item, $p$ refers to the output information of the pre-trained model, $q$ is the label, $\alpha, \beta$, and $\delta$ are three hyper-parameters. The loss function also uses the label information and model output. 
	\subsubsection{Ensemble Method}
	Once obtaining the pre-trained model, we need to maximize the advantages of each model. So, we use ensemble each model with the Bagging Algorithm \cite{skurichina2002bagging} via voting on the predicted results of the trained models. The Bagging algorithm is used during the prediction, where this method can effectively reduce the variance of the final prediction by bridging the prediction bias of different models, enhancing the overall generalization ability of the system.
		\begin{table}[t]
		\centering
		\renewcommand\arraystretch{1.2}	\setlength{\tabcolsep}{2.7mm}
		
		\begin{tabular}{c|ccc|c}
			\noalign{\hrule height 1pt}
			Model     & FGM  & ChildTune  & In-trust loss & Backbone\\ \noalign{\hrule height 0.5pt}
			DeBERTa-large  & 61.27    &61.23   & 61.54   &61.42   \\ 
			mDeBERTa-base &61.89 & 61.68 & \textbf{62.21} & 62.06\\
			\noalign{\hrule height 1pt}
		\end{tabular}
		\caption{The experimental results of the text detection.}
				\vspace{-0.7cm}
		\label{sci}
	\end{table}
	\section{Experiment}
	We will introduce the RuATD dataset, evaluation indicators, implementation details and method description.
	\subsection{RuATD}
	
	However, some people may use these models with malicious intent to generate false news, automatic product reviews, and false political content. RuATD 2022 proposes a new task, which requires judging whether a sentence is generated by the model (binary classification) or even which model it is generated by (multi-class classification). More task details can be found in the website\footnote{\url{https://www.kaggle.com/c/ruatd-2022-multi-task}}.
	
	\subsection{Evaluation}
	In the multi-classification task, accuracy is used as the evaluation index. The task requires the model to judge whether a sample is written by humans or generated by other generation models.
	\begin{equation}
		Acc=\big(\frac{Right}{All})\nonumber
	\end{equation}

	\subsection{Baseline Introduction}\par
	\textbf{Tf-idf:} With the help of sklearn\cite{pedregosa2018scikitlearn}, the organizers connect TF-IDF, SVD, standardscaler and logistic regression in turn for training.\par
	\textbf{ BERT Fine-tuning:}
	BERT\cite{devlin2018bert} is a model designed for natural language understanding task. It uses MLM pre-training method and has strong semantic feature understanding ability.The organizer added a 14 category linear layer after the output layer of BERT-base, and Cross-Entropy loss is used to fit.
	
	\subsection{Implementation Details}
	
	We train the model using the Pytorch \footnote{\url{https://pytorch.org}} \cite{NEURIPS2019_bdbca288} on the NVIDIA RTX3090 GPU and use the hugging-face\footnote{\url{https://github.com/huggingface/transformers}} \cite{wolf-etal-2020-transformers} framework. For all uninitialized layers, We set the dimension of all the hidden layers in the model as 1024.  The AdamW\cite{IlyaLoshchilov2018DecoupledWD} optimizer which is a fixed version of Adam \cite{DiederikPKingma2014AdamAM} with weight decay, and set $\beta_1$ to 0.9, $\beta_2$ to 0.99 for the optimizer. We set the learning rate to $1e-5$ with the warm-up \cite{7780459}. The batch size is 32. We set the maximum length of $280$, and delete the excess. Linear decay of learning rate and gradient clipping is set to $1e-4$. Dropout \cite{NitishSrivastava2014DropoutAS} of $0.1$ is applied to prevent over-fitting. All experiments select the best parameters in the valid set. Finally, we report the score of the best model (valid set) in the test set.
	
	We use the mDeBERTa-base \cite{he2021deberta,PengchengHe2021DeBERTaV3ID} as our pretrained model, and fine-tune the model \footnote{You can reproduce the baseline code from here \url{https://github.com/dialogue-evaluation/RuATD/blob/main/Baseline.ipynb}}. The mDeBERTa\footnote{\url{microsoft/mdeberta-v3-base}} model comes with 12 layers and a hidden size of 768. And it was trained with the CC100 \cite{conneau-etal-2020-unsupervised} multilingual data .
		\begin{table}[t]
	\centering
	\renewcommand\arraystretch{1.2}	\setlength{\tabcolsep}{7.5mm}
	
	\begin{tabular}{cc}
		\noalign{\hrule height 1pt}
		Methods     & Accuracy \\ 
				\noalign{\hrule height 0.5pt}
		Random sample & 19.927 \\
		Tf-idf  & 44.280
		\\ 
		BERT fine-tuning &59.813 \\
		\hline
		
		\textbf{Ours} & \textbf{64.731} 
		\\
		\noalign{\hrule height 1pt}
	\end{tabular}
	\caption{Comparison with baselines in official test set.}
	\label{sc2i}
\end{table}
	\begin{figure}[t]
	\centering
	\includegraphics[scale=0.45]{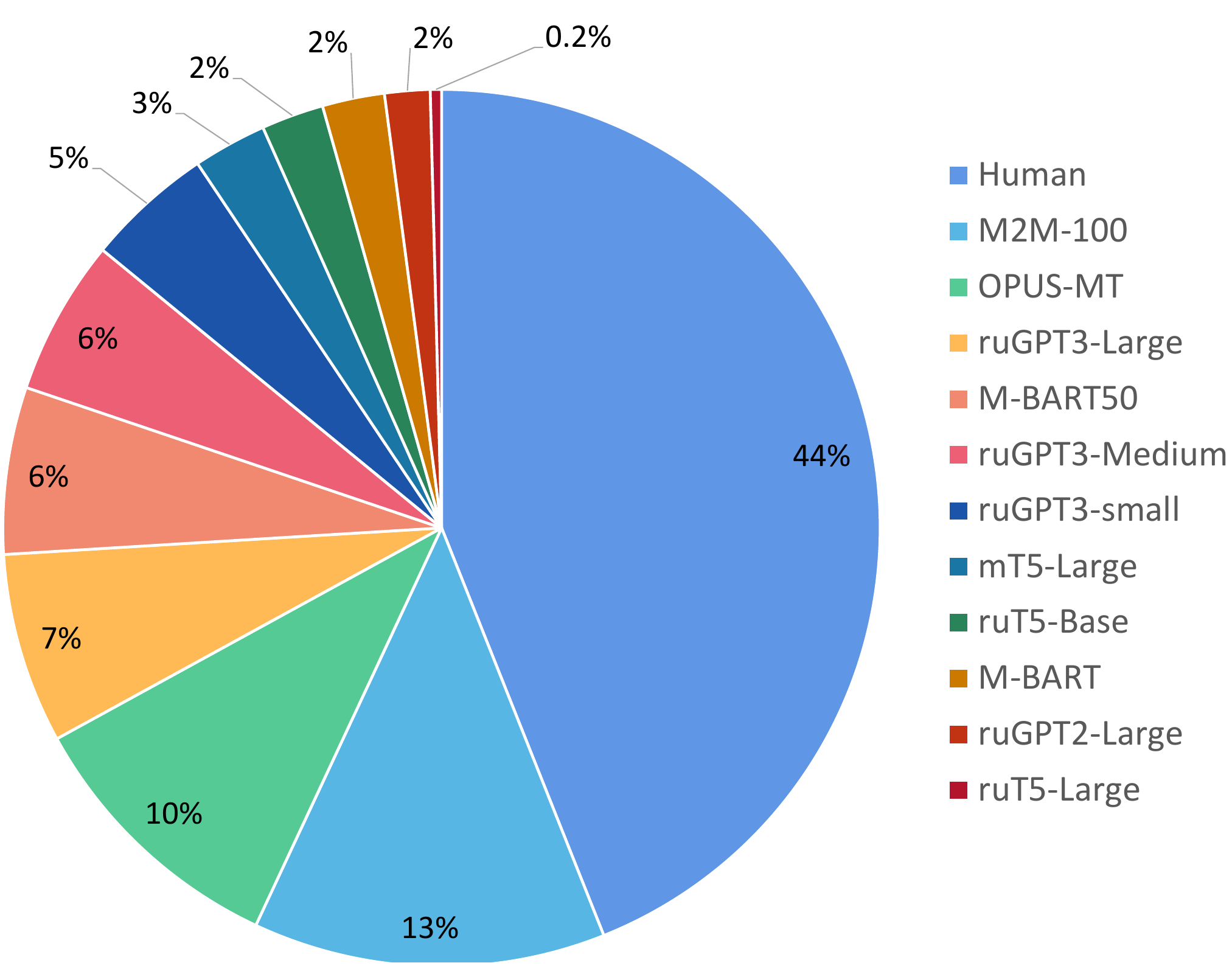}
	\caption{Case study in the shared task.}
			\vspace{-0.5cm}
	\label{fig3}
	\end{figure}
	\section{Case Study}
	We counted and analyzed the mispredicted samples, and the distribution of error types is shown in Figures \ref{fig3}. We chose the top 100 samples with the most significant difference from the ground truth as the analysis object.
	As we can see from Figure \ref{fig3}, the most mispredicted type in the classification task was ``Human", with 44\%, followed by``M2M-100" with 13\%, etc, the ``ruT5-Large'' obtains the least error with 0.2\%.  Further conclusions can be that sorting from high to low actually shows the capability performance of the model. The higher the error rate, the better the performance of the model, and the effects like M2M/GPT3 are better. Then the bigger the model, the harder the target is to distinguish.
	\section{Result and Discussion}
	As shown in Table \ref{sci}, we implement the DeBERTa-large and mDeBERTa-large with multiple training strategies. It can be further concluded that the in trust loss method with the pre-trained model can achieve the best results in artificial text detection. It may be the reason that the model is trained through In-trust training. can be more robust. Moreover, we found that the mDeBERTa outperforms the original version, which indicates that the multi-lingual can provide differentiated knowledge for this text detection. Table \ref{sc2i} also presents the comparison between ours and baselines, where our method outperforms the BERT baseline by 2.397 in accuracy score on the official test set.
	\section{Conclusion}
	This paper illustrates our contributions for Russian Artificial Text Detection Dialogue Shared task (Multi Class). We use the DeBERTa pre-trained language model with multiple traning strategies to distinguish which model from the list was used to generate this text. In the evaluation phase, our submission achieves \textbf{second place}.

	
	\bibliography{dialogue.bib}
	\bibliographystyle{dialogue}
	
	
	
\end{document}